\documentclass[conference,a4paper]{IEEEtran}

\ifCLASSINFOpdf
  \usepackage[pdftex]{graphicx}
  \graphicspath{{../figures/}}
  \DeclareGraphicsExtensions{.pdf,.jpeg,.png}
\fi

\usepackage{amssymb}
\usepackage{graphicx}
\usepackage[export]{adjustbox}
\usepackage{amsmath,amssymb} 
\usepackage[colorlinks]{hyperref}
\usepackage{cleveref}
\usepackage{array}
\usepackage{caption}
\usepackage{subcaption}
\usepackage{multirow}

\usepackage{color}
\usepackage{array}
\usepackage{authblk}

\newcolumntype{M}[1]{>{\centering\arraybackslash}m{#1}}

\begin{document}
\title{Picture What you Read}

\author{Ignazio Gallo}
\author{Shah Nawaz}
\author{Alessandro Calefati}
\author{Riccardo La Grassa}
\author{Nicola Landro}

\affil[1]{Department of Theoretical and Applied Science, University of Insubria, Varese, Italy}
\affil[ ]{\tt\small {\{ignazio.gallo,snawaz,a.calefati,rlagrassa\}@uninsubria.it}}



\maketitle
\begin{abstract}
Visualization refers to our ability to create an image in our head based on the text we read or the words we hear. 
It is one of the many skills that makes reading comprehension possible. 
Convolutional Neural Networks (CNN) are an excellent tool for recognizing and classifying text documents. In addition, it can generate images conditioned on natural language. 
In this work, we  utilize CNNs capabilities to generate realistic images representative of the text illustrating the semantic concept.
We conducted various experiments to highlight the capacity of the proposed model to generate representative images of the text descriptions used as input to the proposed model.
\end{abstract}

\section{Introduction}
Recent years have seen a surge in multimodal data containing various media types. Typically, users combine text, image, audio or video to sell a product over an e-commence platform or express views on social media. The combination of these media types has been extensively studied to solve various tasks including classification~\cite{arevalo2017gated,kiela2018efficient,gallo2017multimodal}, cross-modal retrieval~\cite{wang2016learning} semantic relatedness~\cite{kiela2014learning,leong2011going}, image captioning~\cite{karpathy2015deep,xu2015show}, multimodal named entity recognition~\cite{zhang2018adaptive,arshad2019aiding} and Visual Question Answering~\cite{fukui2016multimodal,anderson2018bottom}. In addition, multimodal data fueled an increased interest in generating images conditioned on natural language~\cite{reed2016generative,cha2017adversarial}.
In recent years, generative models based on conditional Generative Adversarial Network (GAN) have remarkably improved text to image generation task~\cite{hong2018inferring,zhu2018generative}. Furthermore, generative models based on Variational Autoencoders are employed to generate images conditioned on natural language~\cite{mansimov16_text2image,mansimov2015generating}. Generally, image generation from natural language is divided into phases: the first phase learns the distribution from which the images are to be generated while the second phase learns a generator, which in turn produces the image conditioned on a vector from this distribution.

\begin{figure}[t]
  \centering 
  \includegraphics[width=0.90\columnwidth]{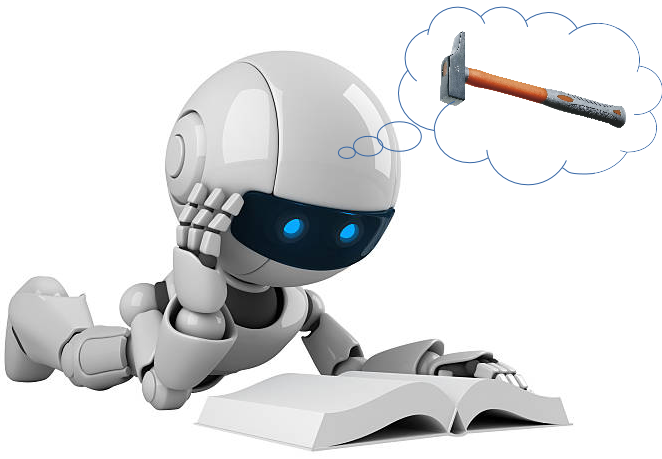}
  \caption{A Neural Model reading natural language (``Heavy Duty All Purpose Hammer - Forged Carbon Steel Head'') can generates a representative image (``Hammer'').}
  \label{fig:idea}
\end{figure}

In this work, we are interested in transforming natural language in the form of technical e-commerce product specifications directly into image pixels. For example, image pixels may be generated from the text description such as ``Heavy Duty All Purpose Hammer - Forged Carbon Steel Head'' as shown in Fig.~\ref{fig:idea}.  We assume we are given technical specifications of a set of images available on e-commence platforms, and train the generator block, available inside our model, from the pixel distribution. 
We propose to use an `up-convolutional' generative block for this task and show that it is capable of generating realistic e-commerce images. 
Fig.~\ref{fig:correct-results} shows some generated images conditioned on technical product specification along with the original images. 
Following are the main contributions of our work:
\begin{itemize}
\itemsep0em 
\item[--] We propose a new loss function to transform a text description into a representative image;
\item[--] The proposed model generates images conditioned on technical e-commerce specifications. Moreover, it generates images never seen before;
\item[--] An end-to-end convolutional model capable of classifying the text and at the same time generating a representative image of the text. The generated image can be used as text encoding or as a realistic image representing the object described in the input document;
\item[--] We propose a model that can also be used to transform a multimodal dataset into a single dataset of images.
\end{itemize}

\begin{figure*}
  \centering 
  \includegraphics[width=0.90\textwidth]{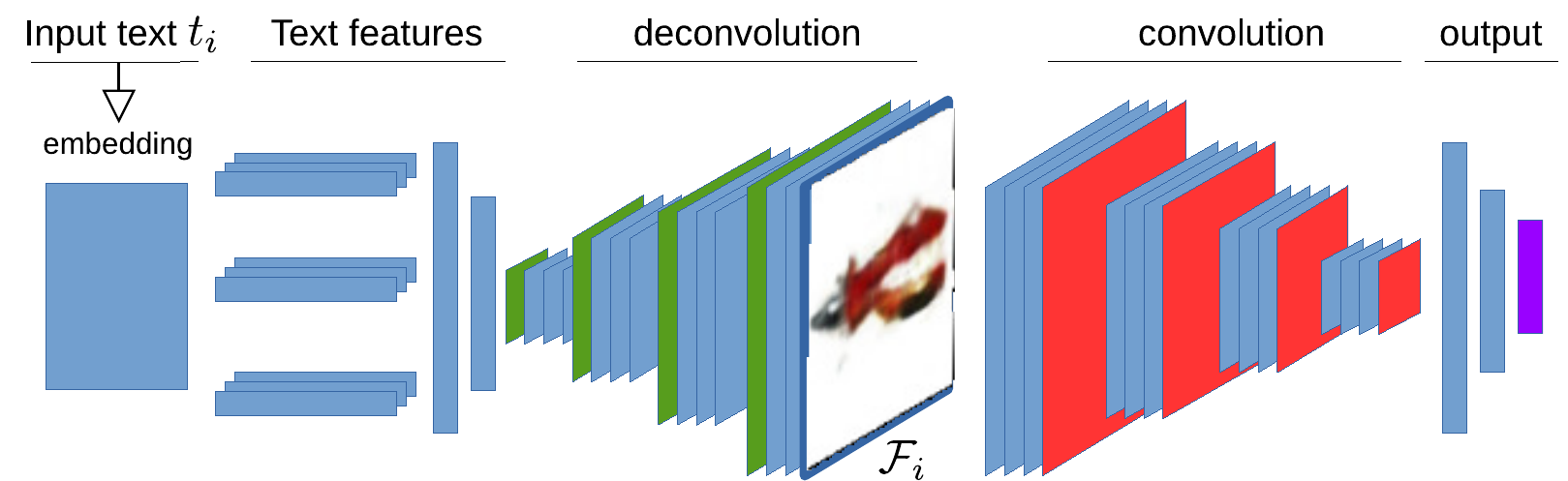}
  \caption{A schematic representation of the proposed model. 
  The model extracts features from the text document using an embedding layer and three different convolutive filters. Through different deconvolutive layers the textual features are transformed into an image representative  of the text. Finally, through convolutive layers, the image and the encoded text are classified.
  }
  \label{fig:schematic}
\end{figure*}

\section{Related Work}
Recently, image generation conditioned on natural language has drawn a lot of attention from the research community. Various approaches have been proposed based on Variational Autoencoders~\cite{mansimov16_text2image,mansimov2015generating}, Auto-regressive models~\cite{reed2017parallel} and optimization techniques~\cite{nguyen2017plug}.
Similarly, GANs based approaches have noticeably improved image synthesis conditioned on natural language. 
These approaches consist of a generator and a discriminator that compete in a two player minimax game: the discriminator tries to distinguish real data from generated images, and the generator tries to trick the discriminator. 
In the proposed model, the generator and discriminator are part of the same model and are linked by a single loss function, in order to generate images within the classification process.
In the following paragraph, we reviewed couple of the ground breaking approaches on image generation conditioned on natural language.

Reed et al.~\cite{reed2016generative} proposed to learn both generator and discriminator conditioned on captions.  Zhu et al.~\cite{zhu2018generative} proposed a generative method to generate synthesized visual features using the noisy text descriptions about an unseen class. Xu et al.~\cite{xu2018attngan} proposed attentional generative network to synthesize fine-grained details at different subregions of the image by paying attentions to the relevant words in the natural language description. Zhang et al.~\cite{zhang2017stackgan} decomposed text to image generation in two stages: the first stage GAN sketches the basic shape and color of the object condition on the natural language, resulting in low resolution image. While the second stage GAN takes
first stage results and natural language to generate high-resolution images with photo-realistic details.

Furthermore, various approaches have exploited the capability of `up-convolutional' network to generate realistic images. Dosovitskiy et al. ~\cite{dosovitskiy2016learning} trained a deconvolutional network with several layers of convolution and upsampling to generate 3D chair renderings given object style, viewpoint and color. 
In this work we are interested in generating new images through up-sampling but we limit this generative process to a medium resolution.

\section{Model description}
\label{proposed-approach}
Our goal is to train a neural network to generate accurate e-commerce images from a low-level and noisy text description.
We develop an effective loss function to transform a text document into a representative image and at the same time exploits the information content of the image and the text, to solve a classification problem.

Formally, we assume that we are given a dataset of examples 
$D = \{t_1,\dots,t_N\}$ with targets $O=\{(y_1, \mathcal{I}_1),\dots,(y_N, \mathcal{I}_N)\}$. 
The inputs $t_i$ are text descriptions describing the objects showed in the images $\mathcal{I}_i$. 
The targets are tuples consisting of two elements: the class label $y_i$ in one-hot encoding and a $RGB$ image $\mathcal{I}_i$.

\begin{figure}
  \centering 
\begin{tabular}{llp{4cm}}
Generated & Original & Input Text   \\ \hline \hline
\includegraphics[valign=T,width=0.20\columnwidth]{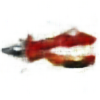} & \includegraphics[valign=T,width=0.20\columnwidth]{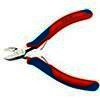} & draper expert knipex 27723 side trimmer for electronics for cutting head satin without facet 115 mm \\ \hline
\includegraphics[valign=T,width=0.20\columnwidth]{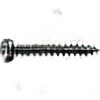} & \includegraphics[valign=T,width=0.20\columnwidth]{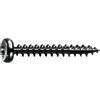} & spax universal screw half round head t star plus 4 cut partial shiny thread galvanized with a2j galvanization   \\ \hline
\includegraphics[valign=T,width=0.20\columnwidth]{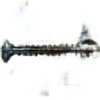} & \includegraphics[valign=T,width=0.20\columnwidth]{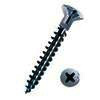} & screws mustad panel vitals bronzed 3x20 mm. conf. 500 pcs universal countersunk flat head screw suitable for screwing ... \\ \hline
\includegraphics[valign=T,width=0.20\columnwidth]{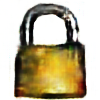} & \includegraphics[valign=T,width=0.20\columnwidth]{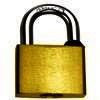} & sicutool padlocks 2090p 50 width body 50 mm \\ \hline
\includegraphics[valign=T,width=0.20\columnwidth]{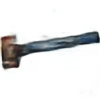} & \includegraphics[valign=T,width=0.20\columnwidth]{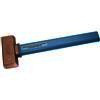} & sicutool copper hammers 2718 800 total weight 800 copper hammers \\ \hline
\includegraphics[valign=T,width=0.20\columnwidth]{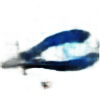} & \includegraphics[valign=T,width=0.20\columnwidth]{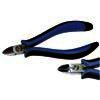} & sicutool nippers for electronics and fine mechanics 557gf lenght total mm 125 wire cutters for electronics and fine mechanics \\ \hline
\includegraphics[valign=T,width=0.20\columnwidth]{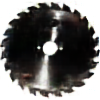} & \includegraphics[valign=T,width=0.20\columnwidth]{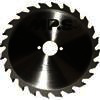} & sicutool circular saws with teeth shown in hartmetall 4840g 150b type null 150b ø mm 150 thickness mm 2 8 hole mm 16 teeth nr 20 \\ \hline
\includegraphics[valign=T,width=0.20\columnwidth]{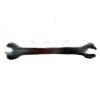} & \includegraphics[valign=T,width=0.20\columnwidth]{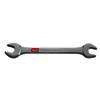} & valex wrench 20 x 22mm inclined forks of 15 ° body in chrome vanadium steel with polished finish. size 20x22 mm length 235 mm \\ \hline
\includegraphics[valign=T,width=0.20\columnwidth]{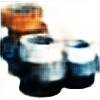} & \includegraphics[valign=T,width=0.20\columnwidth]{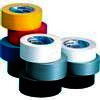} & 
syrom adhesive tape in textile tes special 38 mm x 2 7 m black tightly woven plasticized tape to repair binding edges etc. \\ \hline
\includegraphics[valign=T,width=0.20\columnwidth]{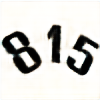} & \includegraphics[valign=T,width=0.20\columnwidth]{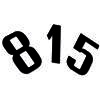} & 
oem 20 pcs sticker number 6 mm 50 black pvc sticker number 6  \\ \hline
\end{tabular}
  \caption{Some examples of generated images from test dataset, associated with a correct classification. In some cases, the generated images are slightly different from the respective expected image but the object shown is very similar.
  }
  \label{fig:correct-results}
\end{figure}

Neural networks typically produce class probabilities by using a ``softmax'' output layer that converts the logit, $a_i$, computed for each class into a probability $p_i$, by comparing $a_i$ with the other logits.
Softmax function takes an $n$-dimensional vector of real numbers and transforms it into a vector of real number in range $[0,1]$ which add upto 1. 
\begin{equation}
  p_i = \frac{\exp(a_i )}{\sum_j \exp(a_j )}
\end{equation}
``Cross entropy'' indicates the distance between what the model believes the output distribution should be ($y_i$), and what the original distribution really is. 
It is defined as
\begin{equation}
\mathcal{L}_0(y, p)=-\sum_{i} y_{i} \log \left(p_{i}\right)
\end{equation}
Cross entropy measure is a widely used alternative of squared error.
If we want to minimize the pixel-by-pixel distance between the input image $\mathcal{I}_i$, associated with the text document and a CNN's features layer $\mathcal{F}_i$ that has the same dimensions as the image $\mathcal{I}_i$, then we can apply the following formula
\begin{equation}\label{eq:L1_loss}
  \mathcal{L}_1(\mathcal{F},\mathcal{I}) = \sum_i (\mathcal{F}_{i} - \mathcal{I}_{i})^2
\end{equation}
or the following mean version
\begin{equation}\label{eq:L1_loss_mean}
  \hat{\mathcal{L}_1}(\mathcal{F},\mathcal{I}) = \frac{1}{N}\sum_i (\mathcal{F}_{i} - \mathcal{I}_{i})^2
\end{equation}
$\mathcal{F}_i$ is the output of the last transposed convolutions -- also called fractionally strided convolutions -- used to upsampling the text features to attain a feature layer having the same size of the image $\mathcal{I}$.

The final loss function we used in this work is the following
\begin{equation}\label{eq:our_loss_function}
  \mathcal{L} = \mathcal{L}_0(y,p) + \lambda \mathcal{L}_1(\mathcal{F},\mathcal{I})
\end{equation}
but we also performed experiments replacing the $\mathcal{L}_1$ with the $\hat{\mathcal{L}_1}$. 
\begin{equation}\label{eq:our_loss_function_mean}
  \hat{\mathcal{L}} = \mathcal{L}_0(y,p) + \lambda \hat{\mathcal{L}_1}(\mathcal{F},\mathcal{I})
\end{equation}
In this last case the contribution of the lambda parameter has a different effect with the same $\lambda$ values, this because the interval of variability of $\mathcal{L}_1$ and $\hat{\mathcal{L}_1}$ are very different.
For example, using the Eq.~\ref{eq:our_loss_function_mean} as loss function, we need much larger $\lambda$ values to take advantage of the contribution of $\hat{\mathcal{L}_1}$ and obtain in $\mathcal{F}$ a realistic image, representative of the object described in the text $t_i$.

The $\lambda$ parameter is important to balance the contribution of $\mathcal{L}_0$ against $\mathcal{L}_1$. Setting $\lambda=0$ we minimize only $\mathcal{L}_0$ and therefore the feature layer $\mathcal{F}$, representing the input text, will be very different from the image $\mathcal{I}$. In Fig.~\ref{fig:generated-image-lamda} we have a graphical representation of the image learned by minimizing Eq.~\ref{eq:our_loss_function}, using various $\lambda$ values and it is important to note that when $\lambda = 0$ the model does not generate realistic images.

Learning proceeds by minimizing the loss function $\mathcal{L}$ via Adam optimizer~\cite{kingma2014adam}. Since we are using the combination of two loss functions to modify the weights of the entire neural network, we know that within the image $\mathcal{F}$ that we generate, it contains text representations.
In fact, in many cases it may happen that the $\mathcal{F}$ image cannot be interpreted as one of the objects belonging to a particular class but the classification is correct.

We experimented with a network for generating images of size $100\times100$. 
The structure of the generative network is shown in Fig.~\ref{fig:schematic}.
Conceptually, the network we propose can be seen as a convolutive classification model for text documents, that incorporates a generative network.
The starting point of the proposed model is the classification model of sentences proposed by Kim~\cite{kim2014convolutional} to transform a text document into a features vector. This is followed by a set of 4 deconvolutive layers that transform textual features into an RGB image that we can then generate. Finally, we used a sequence of 4 convolutive layers that transform the generated image into features for the classification problem.

\subsection{Text features}
The input to our model are sequences of words $[w_i,\dots , w_{|t|}]$ from each input document $t$, where each word is drawn from a vocabulary $V$.
Words are represented by distributional vectors $\mathbf{w} \in \mathbb{R}^{1\times d}$ looked up in a word embeddings matrix $\mathbf{W} \in \mathbb{R}^{d\times |V|}$. 
This matrix is formed by simply concatenating embeddings of all words in $V$.

For each input text $t$, we build a matrix $\mathbf{S} \in \mathbb{R}^ {d\times |t|}$, where each row $i$ represents a word embedding $w_i$ at the corresponding position $i$ in the document $t$. 
To capture and compose features of individual words in a given text from low-level word embeddings into higher level semantic concepts, the neural network applies a series of transformations to the input matrix $S$ using convolution, non-linearity and pooling operations.
The convolution operation between $S$ and a filter $\mathbf{F} \in \mathbb{R}^{d\times m}$ of height $m$ results in a vector $\mathbf{c} \in \mathbb{R}^{|t|+m−1}$. 
In our model we used three groups of $128$ different kernels in parallel, having dimensions $d\times 3$, $d\times 4$ and $d\times 5$.
In this way we obtained three feature maps $\mathbf{c}_i$, having different lengths.
Note that the convolution filter is of the same dimensionality $d=128$ as the input sentence matrix, so this is like a 1-D convolution operation.
To allow the network to learn an appropriate threshold, we also added a bias vector $b \in \mathbb{R}^n$ for each feature map $\mathbf{c}_i$.

Each convolutional layer is followed by the Rectified Linear Unit (ReLU) non-linear activation function, applied element-wise.
ReLU speeds up the training process~\cite{nair2010rectified}, defined as $\max(0, x)$ to ensure that feature maps are always positive.
To capture the most important feature -- one with the highest value -- for
each feature map, we apply a max-overtime pooling operation~\cite{collobert2011natural} over each feature map.
In this way, for each  particular filter we take the maximum value as the feature corresponding to this particular filter.
The convolutional layer utilizing the activation function and the
pooling layer acts as a non-linear feature extractor.

Up to this point we have described the process by which a single feature is extracted from a single filter.
The set of these individual features are linked into a single layer and then connected to a subsequent fully-connected layer that has the purpose of connecting the textual features with the next block of deconvolution layers used to transform textual features into an image.

\subsection{Up-sampling}
The purpose of the up-sampling block is to transform the features extracted from the text into image format that best represents the description contained in the text.

We use 4 deconvolution layers, each of which doubles the size of the input features.
In practice, we start with 512 features maps of size $7\times 7$ and then move to a second layer with 256 features maps of size $13\times 13$, followed by a new layer having 128 features maps o size $25\times25$, another layer with 64 features maps of size $50\times50$ and finally, a last layer $\mathcal{F}$ with 3 features maps of size $100\times100$.
The up-sampling blocks consist of the nearest-neighbor up-sampling followed by a $5\times5$ stride 1D convolution. 
Batch normalization and ReLU activation are applied after every convolution except the last one where we used a sigmoid to guarantee that the values of features maps $\mathcal{F}$  were all in the range $[0,1]$. 
This last layer can be interpreted directly as an image.

\subsection{Classification}
The last block is similar to a convolutive neural network that feeds the features $\mathcal{F}$  to 4 successive convolutive layers.
Each of these layers produces features maps of size $50\times50$, $25\times25$, $13\times13$ and $7\times7$ respectively.
Starting from the largest layer, we used $2\times2$ stride and the following number of $5\times5$ filters: 64, 32, 16 and 8 respectively.

The convolutional layers are followed by two fully connected layers having 1024 and 512 neurons respectively.

For regularization we employ dropout before the output layer with a constraint on $l_2$-norms of the weight vectors. 
Dropout prevents co-adaptation of hidden units by randomly dropping out a proportion $p$ of the hidden units during foward backpropagation.

The last layer is the output layer that has a number of neurons equal to the number of classes of the problem that we want to learn.

\begin{figure}
  \centering 
  \includegraphics[width=1.0\columnwidth]{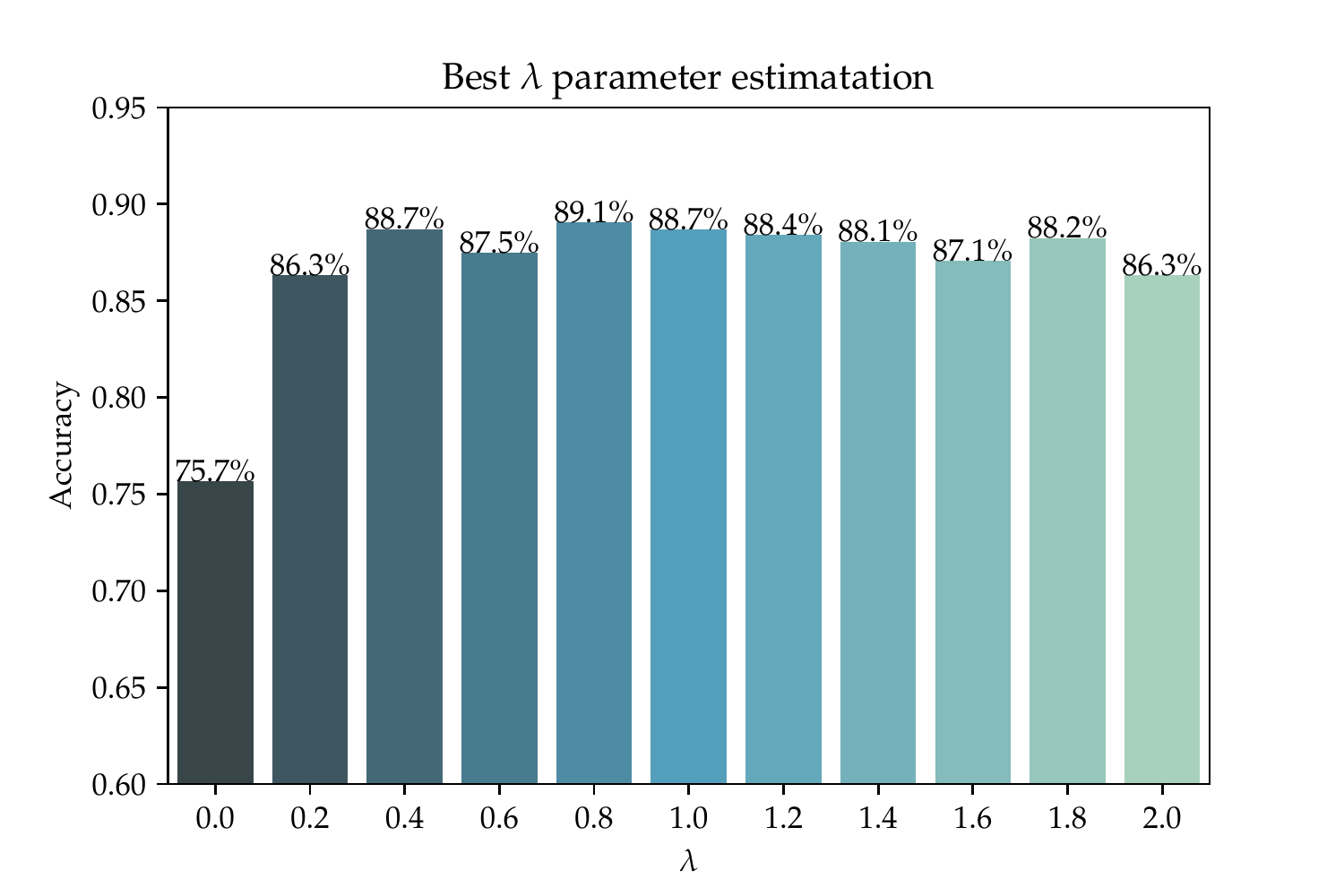}
  \caption{11 different executions made on the validation set of the Ferramenta dataset, to estimate the best value for the $\lambda$ parameter of the proposed loss function in Eq.~\ref{eq:our_loss_function}. 
  The best value obtained is for $\lambda = 0.8$.
  }
  \label{fig:best-lambda}
\end{figure}

\section{Datasets}
In multimodal dataset, modalities are obtained from multiple input sources.
Dataset used in this work consists of images and accompanying text descriptions.
We select Ferramenta~\cite{gallo2017multimodal} multimodal dataset that are created from e-commerce website.
Table~\ref{tab:dataset-info} shows information on this dataset.
Ferramenta multi-modal dataset~\cite{gallo2017multimodal} is made up of $88,010$ adverts split in $66,141$ adverts for train set and $21,869$ adverts for test set, belonging to $52$ classes.
Ferramenta dataset provides a text and a representative image for each commercial advertisement.
It is interesting to note that text descriptions in this dataset are in Italian Language. 

\begin{table}[]
\begin{center}
\caption{Information on multi-modal datasets used in this work. 
A multi-modal dataset consists of an image and accompanying text description. 
The last column indicates the text description language.}
\label{tab:dataset-info}
\begin{tabular}{|lcccc|}
\hline
Dataset & \#Cls & Train  & Test  &  Lang. \\
\hline
Ferramenta   	& 52		& 66,141	& 21,869	& IT \\
\hline
\end{tabular}
\end{center}
\end{table}

Another dataset used in our work is the Oxford-102 Flowers dataset~\cite{nilsback2008automated} containing 8,189 flow images in 102 categories. 
Each image in this dataset is annotated with 10 descriptions provided by~\cite{Reed:2016:Generative}.
Because this dataset has class-disjoint training and test sets, with 82 train+val and 20 test classes, we randomly shuffled all the classes and split back into training and test.
In this way, all the classes available in the training set are also present in the test set.

\begin{figure*}
  \centering 
\begin{tabular}{cccccc}
$\lambda=0$ & $\lambda=0.2$ & $\lambda=0.4$ & $\lambda=0.6$ & $\lambda=0.8$ & $\lambda=1$   \\ \hline \hline
\fbox{\includegraphics[width=0.25\columnwidth]{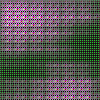}} & 
\fbox{\includegraphics[width=0.25\columnwidth]{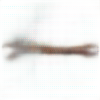}} & 
\fbox{\includegraphics[width=0.25\columnwidth]{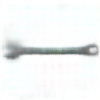}} &
\fbox{\includegraphics[width=0.25\columnwidth]{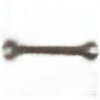}} &
\fbox{\includegraphics[width=0.25\columnwidth]{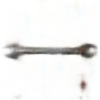}} &
\fbox{\includegraphics[width=0.25\columnwidth]{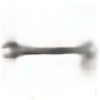}} \\ \hline
$\lambda=1.2$  & $\lambda=1.4$  & $\lambda=1.6$  & $\lambda=1.8$  & $\lambda=2$ & Expected \\ \hline
\fbox{\includegraphics[width=0.25\columnwidth]{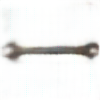}} &
\fbox{\includegraphics[width=0.25\columnwidth]{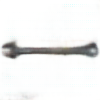}} &
\fbox{\includegraphics[width=0.25\columnwidth]{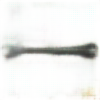}} &
\fbox{\includegraphics[width=0.25\columnwidth]{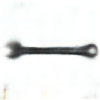}} &
\fbox{\includegraphics[width=0.25\columnwidth]{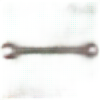}} &
\fbox{\includegraphics[width=0.25\columnwidth]{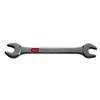}} \\ \hline

\end{tabular}
  \caption{Some examples of generated images from Ferramenta test dataset. 
  The models trained for each $\lambda$ by minimizing Eq.~\ref{eq:our_loss_function}, were trained for 20 epochs only, to speed up the experiment.
  }
  \label{fig:generated-image-lamda}
\end{figure*}

\section{Experiments}

The proposed approach transforms text descriptions into a representative image.
We use standard CNN hyperparameters. The initial learning rate is set to $0.001$ along with Adam as optimizer. 
In our experiments, accuracy is used to measure classification performance.

The purpose of the first experimental phase is to analyze the generative capacity of our model.
We conducted following experiments with this aim in mind: 
(1) estimate of the best $\lambda$ parameter for Eq.~\ref{eq:our_loss_function}, 
(2) qualitative analysis of the generated images $\mathcal{F}$, 
(3) ability of the model to generate new images according to the description given in input.

As a second group of experiments we have analyzed the capacity of the proposed model to generate embedding in image format of the input text.
We conducted following experiments with this second aim in mind: 
(1) estimate of the best lambda parameter to obtain the most significant encoding,
(2) extraction of a new dataset of encoded text in image format to compute the classification accuracy using a well-known CNN.

The first experiment concerned the estimation of the best $\lambda$ value to be used in the proposed loss function described in  Eq.~\ref{eq:our_loss_function}. 
To achieve this, we first extracted a validation set from our training set and on this we calculated the classification accuracy to extract the best value to assign to the lambda parameter.
Fig.~\ref{fig:best-lambda} shows the results of all the experiments conducted on the validation set.
The accuracy results reported in this figure were obtained by averaging the accuracy values of 5 runs.
As can be seen from the figure, the best value we obtained is for $\lambda = 0.8$.
To visually analyze the effect of the $\lambda$ parameter, we also performed a quick test by training 11 different models for 20 epochs using $\lambda \in \{0, 0.2, 0.4, 0.6, 0.8, 1, 1.2, 1.4, 1.6, 1.8, 2\}$.
Then we compared some of the resulting images as shown in Fig.~\ref{fig:generated-image-lamda}.
The best defined image is for $\lambda = 0.8$ while for $\lambda = 0$ we have an abstract visual representation since the second part of the loss function described in Eq.~\ref{eq:L1_loss} has been removed.

As a second experiment, we first trained a model using the entire training set and then visually analyzed generated images with our model on the test set.
Fig.~\ref{fig:correct-results} shows some examples of generated images beside the images we expected and the text passed as input.
As you can see from the figure, many of the images generated are identical to those we expect to find, while some of them represent the same object but arranged differently (see for example the screw and the pliers with the red handle).
We observed that some images have no visual meaning and do not represent any of the objects in the training set, even if in many of these cases the classification is correct.
This means that the information extracted from the text is still present in the image which is then used by the last block to classify.

\begin{figure*}
  \centering 
  \includegraphics[width=1.0\textwidth]{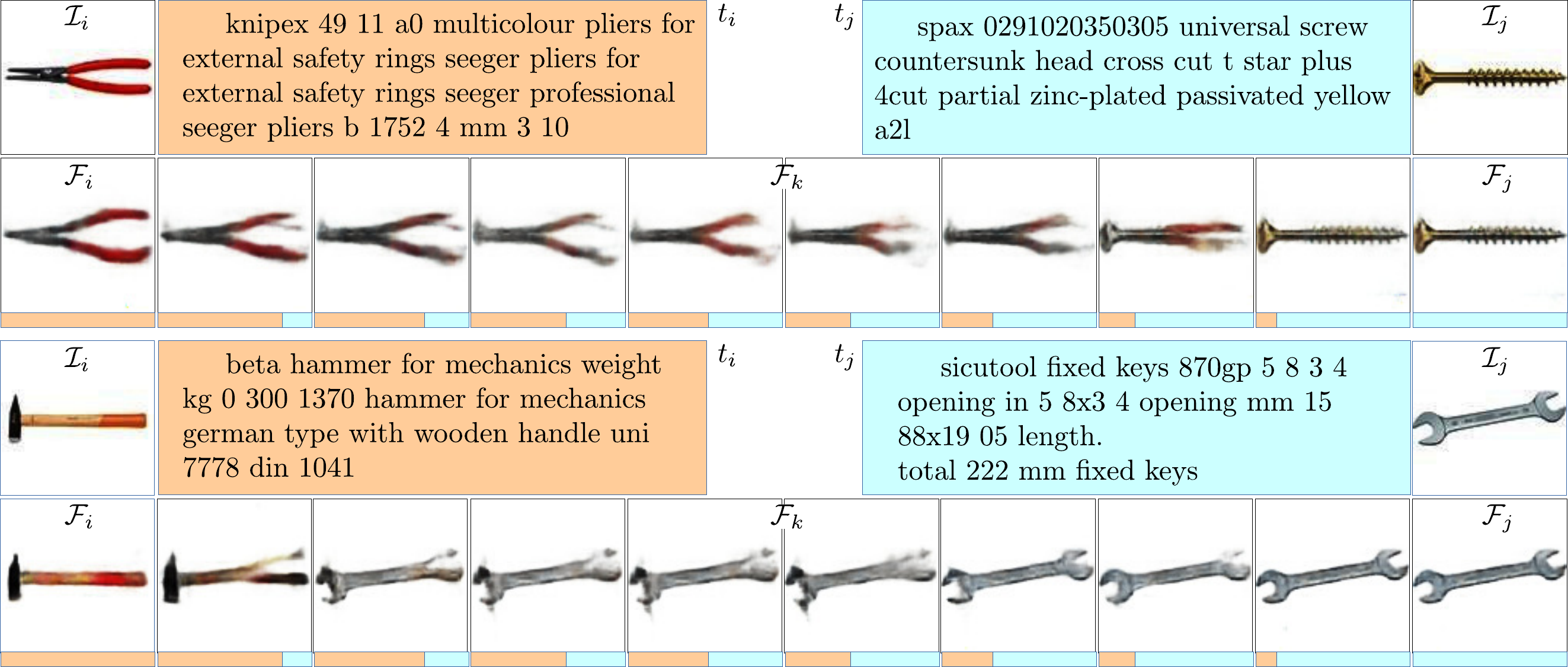}
  \caption{A set of $\mathcal{F}_k$ images generated by the proposed model. 
  $\mathcal{F}_i$ and $\mathcal{F}_j$ have generated starting from text documents $t_i$ and $t_j$ respectively. 
  All other $\mathcal{F}_k$ images were generated by combining different percentages of tokens extracted simultaneously from $t_i$ and $t_j$. 
  The two colored rectangles below $\mathcal{F}_k$ images are indicative of the percentages of tokens from $t_i$ and $t_j$ used to generate the  $\mathcal{F}_k$ images.
  }
  \label{fig:mixed_images}
\end{figure*}

In generalization, the proposed model has the ability to generate images that it has never seen in training and this ability is directly correlated with the words we feed in input.
In this experiment we tried to mix the tokens of two descriptions belonging to different categories to highlight the capacity.
As can be seen from Fig.~\ref{fig:mixed_images}, by taking some tokens from two different descriptions and feeding them to the neural model, in some cases this produces images that are a combination of the objects representing the two descriptions.
For example, in the same Fig.~\ref{fig:mixed_images} you can see an image of an object that is the composition of a screw and a clamp.
This is because in the description given as input to the model, the most important tokens of both objects are present.

\begin{figure}
  \centering 
  \includegraphics[width=0.9\columnwidth]{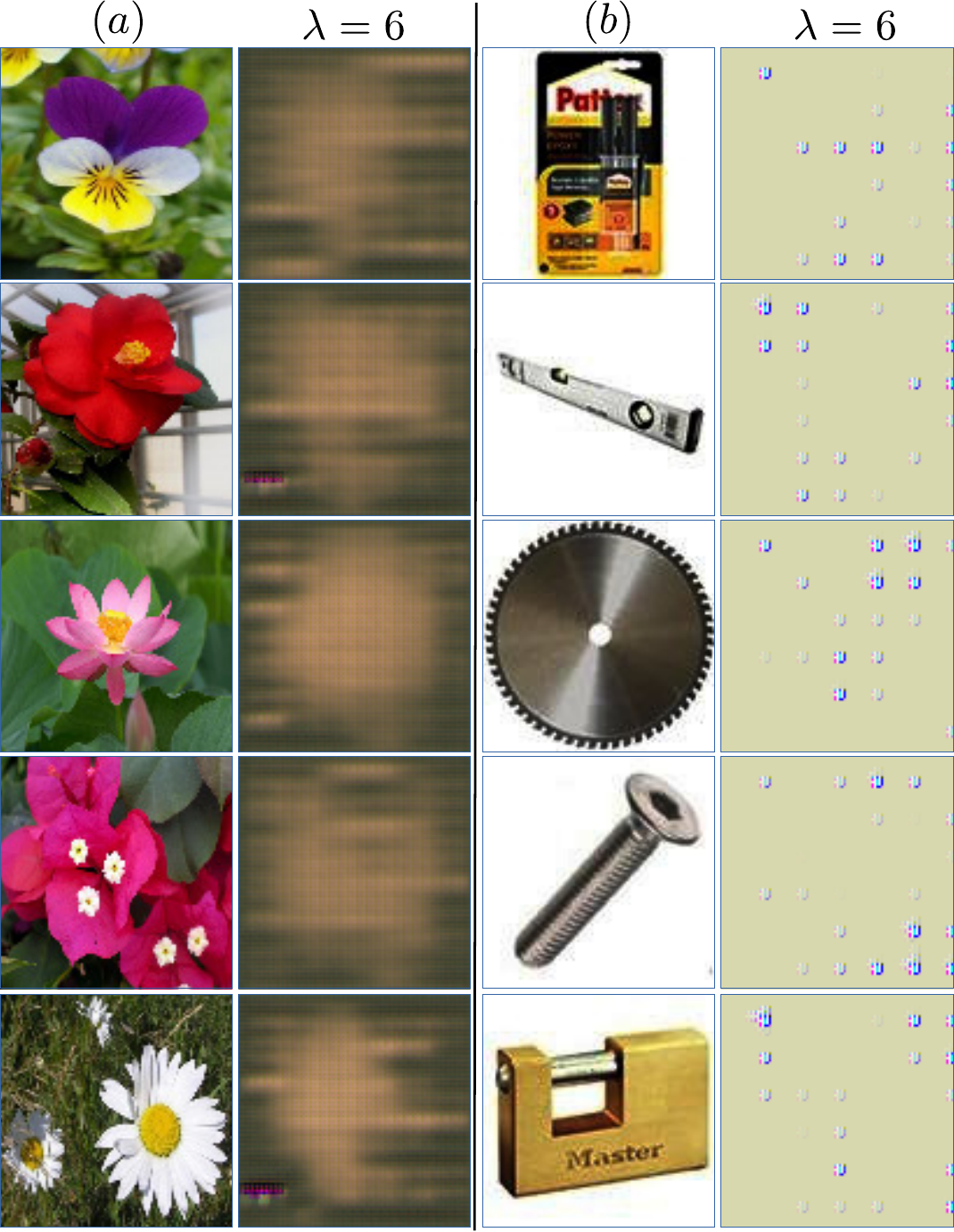}
  \caption{Columns (a) and (b) show some examples of images extracted from Flowers and Ferramenta datasets, respectively. The columns to the right of (a) and (b) show the text encodings extracted as features layer $\mathcal{F}$ using $\lambda=6$ in Eq.~\ref{eq:our_loss_function_mean}.
  }
  \label{fig:encoding-examples}
\end{figure}

\begin{figure}
  \centering 
  \includegraphics[width=1.0\columnwidth]{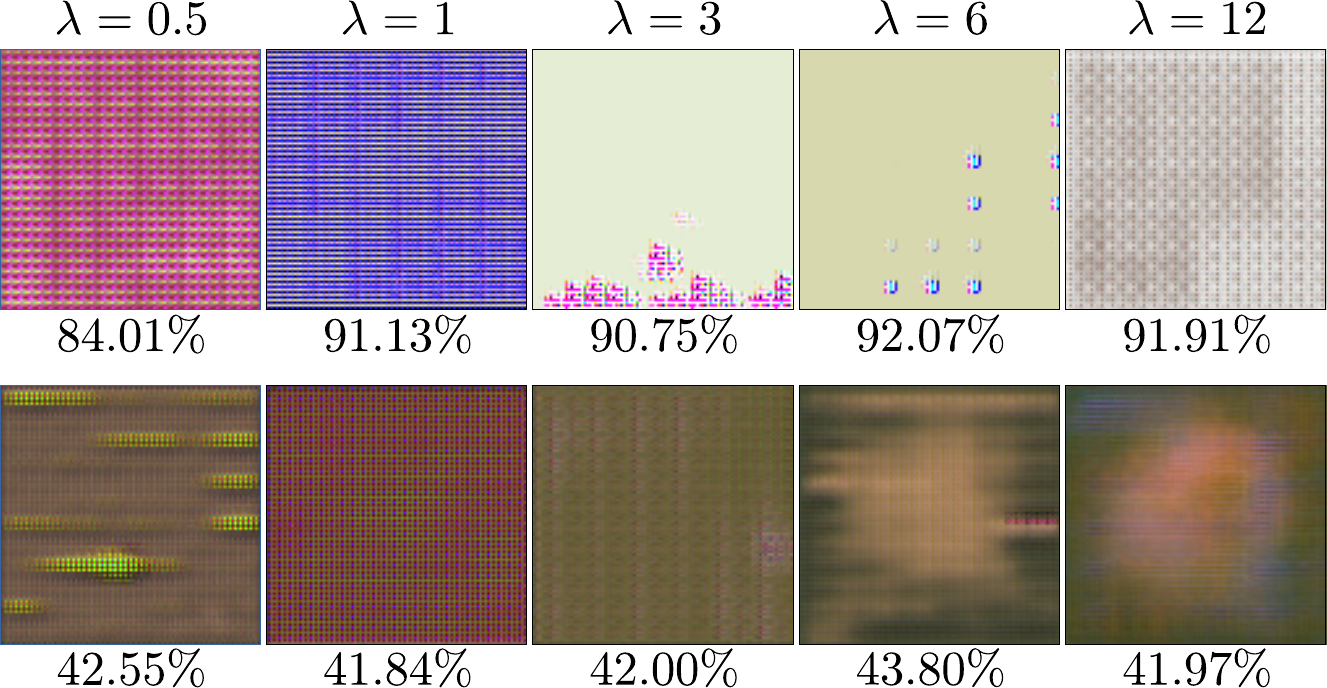}
  \caption{The top row contains encodings for the same image belonging to the Ferramenta dataset when the model was trained using Eq.~\ref{eq:our_loss_function_mean} with $\lambda$ parameters showed on the top. 
  The images on bottom row are created using an image of the Flowers dataset. 
  Below each image is the test accuracy obtained with the corresponding $\lambda$ parameter.
  }
  \label{fig:encoding-varing-lambda}
\end{figure}

Using the loss function of Eq.~\ref{eq:our_loss_function_mean}, which has a much more restricted range of variability, it is possible to give more emphasis to the encoding of the input text in image format.
To find the best $\lambda$ parameter that produces the best encoding we varied the parameter and for each of its values we  computed the classification accuracy on the test set of the Ferramenta and Flowers datasets.
Fig.~\ref{fig:encoding-varing-lambda} shows the encodings and the accuracy obtained when the parameter changes.
On these results we performed the last experiment using the $\lambda = 6$ parameter.

In this latest experiment we extracted two new image datasets using two different models trained on the two datasets.
In Fig.~\ref{fig:encoding-examples} you can see some examples of images generated, alongside the original image of the dataset.
It can be seen how different source images correspond to a different text encoding.
To analyze the information content of these two new datasets we have trained two CNN AlexNet to calculate their classification accuracy.
For the Ferramenta dataset we got $93.68\%$ while for the Flowers dataset we got $99.05\%$.
The first result is slightly higher than the one published in~\cite{Gallo:2018:DICTA} while the second result obtained on the Flowers dataset is incredibly high.
The reason we got such high accuracy is because our dataset contains 10 different text descriptions associated with the same image.
Having divided the training set randomly into training and test, the same images can be found both in the training set and in the test set.
Ultimately this means that the new datasets created do not only encode information extracted from the text but also from images.

\section{Conclusion}
In this work we have proposed a new approach to generate an image that is representative of a noisy text description available in natural language.
The approach we proposed uses a new loss function in order to simultaneously minimize the classification error and the distance between the desired image and a features map of the same model.
The qualitative results are very interesting but, for the moment, we have ignored the classification performances because this was not our focus of the present work.
In the future we want to exploit the same idea to try to improve the classification accuracy that can be obtained with a single convolutive neural model.

Another interesting aspect emerged from this work is that the same approach we proposed can be used to encode in image format both the information contained in the input text and the information extracted from the image associated with the text.
This feature is very interesting to be able to incorporate multimodal information into a single image dataset.
In this way, multimodal information can be processed directly by a single CNN normally used to process only images.

\bibliographystyle{IEEEtran}
\bibliography{bib}

\end{document}